\documentclass{INTERSPEECH2023}

\usepackage{cite}
\usepackage{graphicx}  
\usepackage{float}  
\usepackage{subfigure}  

\interspeechcameraready

\title{Cascaded Multi-task Adaptive Learning Based on Neural Architecture Search}
\name{Yingying Gao, Shilei Zhang, Zihao Cui, Chao Deng, Junlan Feng$^{\ast}$ \thanks{*Corresponding author}}
\address{
  China Mobile Research}
\email{(gaoyingying,zhangshilei,cuizihaoyjy,dengchao,fengjunlan)@chinamobile.com}

\begin{document}

\maketitle
 
\begin{abstract}
Cascading multiple pre-trained models is an effective way to compose an end-to-end system. However, fine-tuning the full cascaded model is parameter and memory inefficient and our observations reveal that only applying adapter modules on cascaded model can not achieve considerable performance as fine-tuning. We propose an automatic and effective adaptive learning method to optimize end-to-end cascaded multi-task models based on Neural Architecture Search (NAS) framework. The candidate adaptive operations on each specific module consist of \textit{frozen}, \textit{inserting an adapter} and \textit{fine-tuning}. We further add a penalty item on the loss to limit the learned structure which takes the amount of trainable parameters into account. The penalty item successfully restrict the searched architecture and the proposed approach is able to search similar tuning scheme with hand-craft, compressing the optimizing parameters to 8.7\% corresponding to full fine-tuning on SLURP with an even better performance.
\end{abstract}

\section{Introduction}

Cascaded systems consist of multiple single-task models concatenated, in which the former output or intermediate layer is delivered to the latter as input. The cascaded structure enables each task to take advantage of the existing well-trained models and data resources, meanwhile makes the upstream and downstream tasks match well. However, the transfer of cascaded model to a particular domain is data and time consuming. Besides, previous research mainly concentrates on single model adaptation, it is still unclear how the transfer mechanism affects the performance in the cascaded multi-task model. This paper focuses on the efficient adaptation for cascaded models.

In recent years, multifarious adapters \cite{gate_adapt,ba,res_adapt,adapt1,adapt2,adapt3,adap4,adapt5} are proposed for a cost-effective transfer which contain a small fraction of model parameters. While these methods concentrate on the compression and the extension of pre-trained models within different adapter modules, the plugging strategy of adapter for different tasks are less studied. \cite{nas_adapt} examined the usage of neural architecture search (NAS) to search for adapter types and plugging locations considering the domain diversity. This is a successful exploration to develop an adaptation scheme automatically, nevertheless, it is still single-task oriented and the cooperation of different tuning approaches is disregarded. In the research about cascaded models, the adaptation of upstream pre-trained models often coordinates with fine-tuning of downstream task to make trade-offs between resource requirements and final performance \cite{pal,pe,kadapter}. \cite{multi-task} proposed a cascaded robust speech recognition model which integrates speech enhancement (SE), self-supervised learning representations (SSLR) and automatic speech recognition (ASR) models. They investigated the benefits of cascaded training of these three tasks and explored efficient training scheme through extensive experiments. \cite{pretrain,pretrain2} introduced adapters to pre-trained self-supervised speech representations and successfully reduced the number of trainable parameters required for downstream ASR tasks. \cite{pretrain} also discussed the possible plugging locations for adapters to gain similar performance to full transfer. \cite{pe} suggested that adapters on the higher-layers had more impacts than lower layers and full fine-tuning is inferior to fine-tuning only the top layers. These approaches obtain optimal adaptation strategy based on artificially designed experiments and explore the cooperation of adapter and fine-tuning by a rather heuristic way.

We aim at designing effective adaptive-tuning scheme for cascaded architectures automatically. In the cascaded multi-task scenario, each component is heterogeneous and the possible plugging position is variable. We also apply NAS to search for appropriate tuning operation in each possible location throughout the cascaded model. The difference with \cite{nas_adapt} is that the candidate operations in our work include \textit{frozen}, \textit{inserting an adapter} and \textit{fine-tuning}, since it has been found that fine-tuning certain layers cooperating with adapter tuning is beneficial or inevitable to achieve optimal performance \cite{residual1,residual2}. The extended fine-tuning operation enables the system to determine the final tuning strategy straightly, superseding hand-craft setting and tedious experiments. To further balance the performance and the parameter amount, we add a penalty on the loss, taking the amount of parameters to be adjusted through each operation into account to guide the learning of architecture parameters in NAS.

The major contributions of this work are threefold:

1) We perform an end-to-end architecture search and optimization for cascaded models by NAS and extend the candidate operations of each possible adapted location into adapting, frozen and fine-tuning.

2) We add a parameter penalty to guide the learning of architecture search in NAS, by which the learned structure size is restricted and the performance is enhanced.

3) The effectiveness of the proposal is verified on a three-task cascaded model, obtains similar tuning scheme with artificial design, decreases the adjusted parameters to 8.7\% and achieves even better performance than full fine-tuning.


\section{Method}

We propose an end-to-end automatic adaptation method for cascaded multi-task adaptive training based on NAS. 
NAS is a search algorithm for automating the design of neural network architectures. 
The basic idea of NAS is to discover optimal architectures from a search space under certain constraints.
The search space is constructed by a list of candidate network operations and their connections, in which the former is known as \textit{network parameters} while the latter is \textit{architecture parameters}.
We build the search space based on multiple cascaded models and the candidate adaptative operations. Each model can be further divided into a series of small modules, and each module corresponds to a possible adapting and searching unit after which an adapter can be inserted or not. The candidate operations for each module in this work include: \textit{frozen}, \textit{inserting an adapter} and \textit{fine-tuning the full module}.

\begin{figure}[h]
  \centering
  \includegraphics[width=\linewidth]{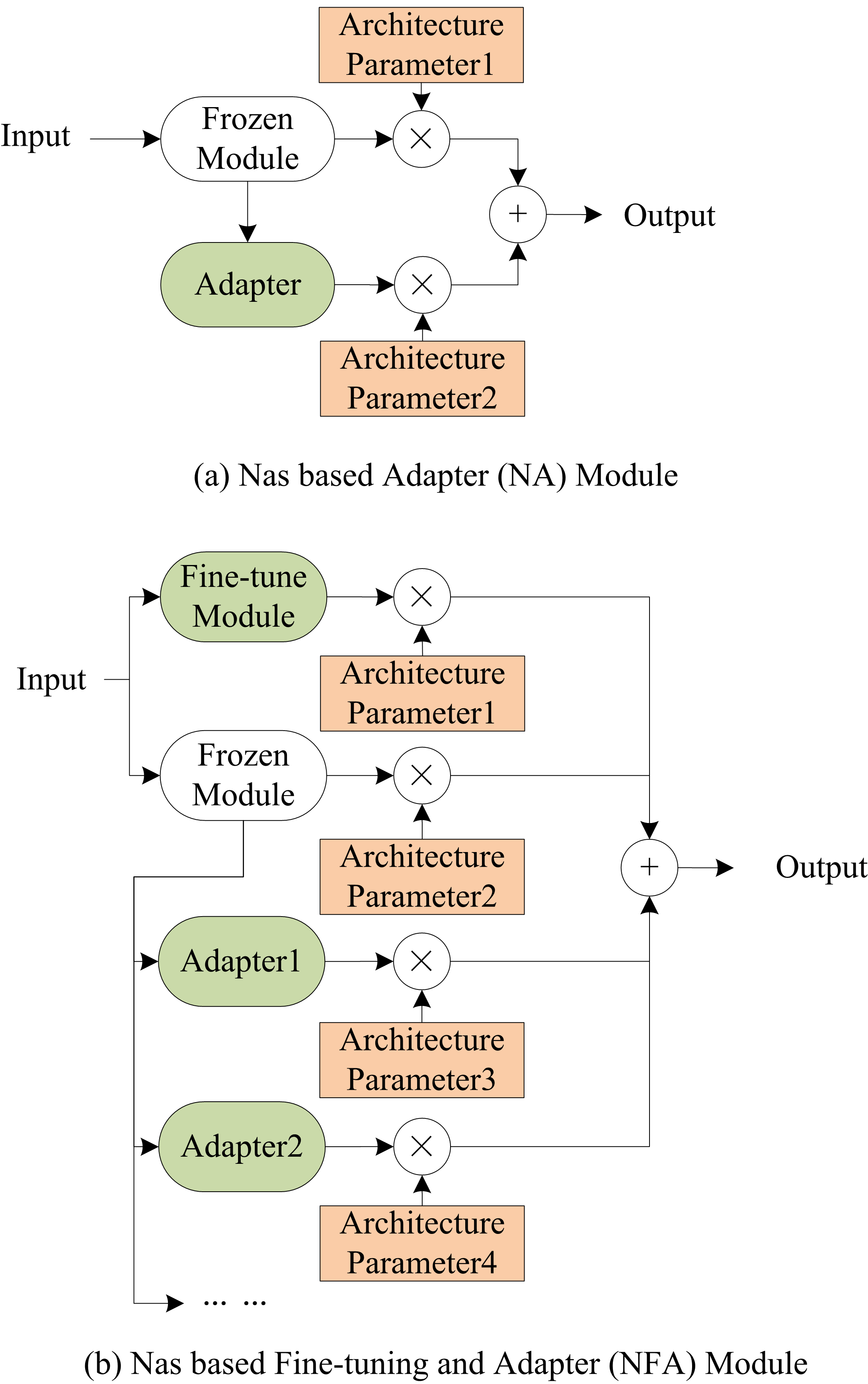}
  \caption{The architecture of NA and NFA module. Only the green and orange boxes are updated during training, in which the green parts denotes trainable network parameters and the orange parts are architecture parameters.}
  \label{fig1}
\end{figure}

\begin{figure}[t]
  \centering
  \includegraphics[width=\linewidth]{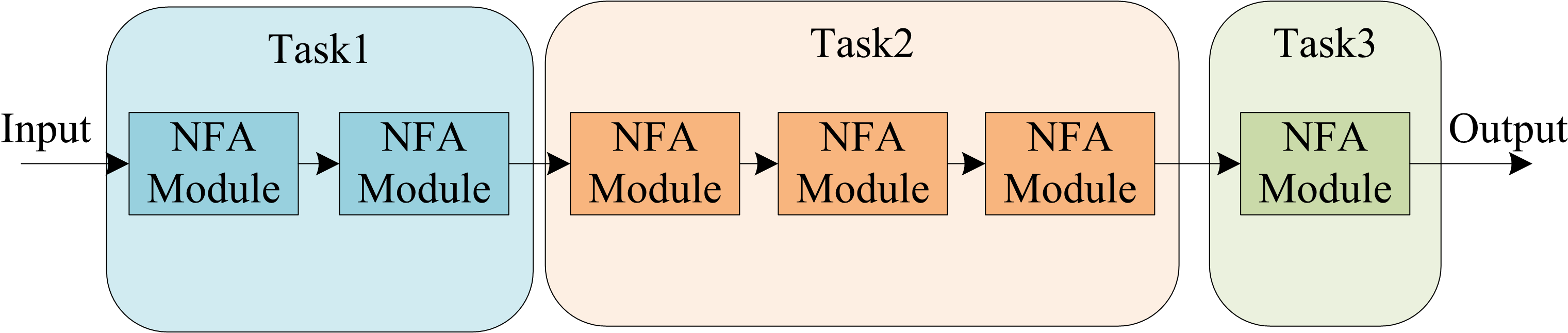}
  \caption{The illustration of an adaptive training framework based on NFA for cascaded multi-task training.}
  \label{fig2}
  \vspace{-0.7cm}
\end{figure}

Figure~\ref{fig1}(a) illustrates the framework of adding an adapter to a module based on NAS, which is named NA in this work. The pre-trained parameters of the module are frozen. After passing the frozen module, two candidate paths are supplied: go through the adapter or not. The connection weights for the candidate paths are set as architecture parameters and the adapter parameters are network parameters. The result of each path is multiplied by the architecture parameter and then summed to get the final output of this module, which is utilized as the unified input of the next module. Figure~\ref{fig1}(b) shows the proposed NFA (Nas based Fine-tuning and Adaptation) module that adds automatic search for whether to fine-tune the module. For each module, three candidate paths are supported: frozen, fine-tuning and adapter. Similarly, the path weights are set as architecture parameters, and the parameters of fine-tuned module and adpter are network parameters.
The final output of the module is the weighted sum of the three paths. The adapter may contain multiple types thus more than one adapter is supplied in Figure~\ref{fig1}(b). Figure~\ref{fig2} demonstrates a multi-task cascaded joint training framework composed of multiple NFA modules connected in series, and different tasks may have different structures and different numbers of NFA modules.

\subsection{Training}

During the training stage, only the adapter or fine-tuned network parameters and architecture parameters are updated. The training data is split into two parts: training set and validation set. In each iteration, the architecture parameters are updated via validation set while the network parameters are frozen; then the adapted or fine-tuned parameters are updated via training set and the architecture parameters remain invariant.

In order to compute a hard architectural weight (0 or 1) for each module and avoid the inconsistency between evaluation and training process, we use Gumbel Softmax \cite{gumble} to calculate the architecture parameter of each path during training, so as to realize path sampling without truncating the gradient.

\subsection{Parameter Control on Loss}

To control the size of learned structure, we add a penalty on the loss to guide the architecture learning. Specifically, the amount of parameters that need to be adapted or fine-tuned on each path is measured then multiplies with the architecture parameter on that path, finally adds up and appends to the raw loss function directly. The motivation of this design is to promote larger modules to choose adapters or to be frozen. As shown in Equation 1, $L_p$ is the proposed loss penalty item related to the trainable parameter amount and architecture parameters, where $i$ is the module index, $M$ is the total number of modules in the cascaded model, $a_{i,ft}$, $a_{i,ad}$ and $a_{i,fr}$ are architecture parameters for fine-tuning, adapter and frozen respectively. $P_{i,ft}$, $P_{i,ad}$ and $P_{i,fr}$ are the number of trainable parameters for each operation, in which $P_{i,ft}$ equals the size of this module, $P_{i,ad}$ is the parameter quantity of the adapter and $P_{i,fr}$ can be set to 0 or a larger number to lower its order. If $P_{i,fr}$ is set as 0 or a small number that is less than $P_{i,ad}$, the final searched result tends to converge to an architecture with plenty of frozen parts while the model performance may be suboptimal. Therefore, we set $P_{i,fr}$ as half of $P_{i,ft}$ , which is in the middle of $P_{i,ad}$ and $P_{i,ft}$, and dynamically changes with different modules. In addition, the weighted summation term is normalized by the sum of trainable parameter numbers of the three paths, to ensure that the parameter penalty item and the original loss are in similar range.

\vspace{-0.7cm}
\begin{align}
  L_{p}=\sum_{i=1}^{M}\frac{a_{i,ft}\cdot P_{i,ft}+a_{i,ad}\cdot P_{i,ad}+a_{i,fr}\cdot P_{i,fr}}{P_{i,ft}+P_{i,ad}+P_{i,fr}}
  \label{equation:eq1}
\end{align}
\vspace{-0.5cm}

\section{Experiments}

The experiments in this work study the cascaded training and adaptive learning of three tasks: speech enhancement (SE), automatic speech recognition (ASR), and natural language understanding (NLU). 

\subsection{Data}

We carry out our experiments on the SLURP corpus \cite{slurp}, which includes speech data, ASR labels (speech content), and semantic tags (intent and slot). It contains 119,820 utterances of training set (84.44 hours in total), 8,679 utterances of validation set (6.87 hours), and 13,060 utterances of test set(10.18 hours). The SE and ASR models are pre-trained by Voicebank \cite{voicebank} and Librispeech \cite{librispeech} respectively (which are directly downloaded from the speechbrain HuggingFace repository\footnote[1]{https://huggingface.co/speechbrain/mtl-mimic-voicebank}$^{,}$\footnote[2]{https://huggingface.co/speechbrain/asr-crdnn-rnnlm-librispeech}), and the NLU model is randomly initialized. We adopt half of SLURP to adapt and fine-tune the cascaded model.

\subsection{Setting}

We based our experiments on the implementation of Voicebank and SLURP recipes in Speechbrain toolkit \cite{speechbrain}. The SE model consists of 8 Transformer blocks as encoder and each has 8 head, 1024 channels and 512 hidden nodes. The candidate adapters of SE model are inserted at the end of each transformer block. The ASR model takes CRDNN as encoder, including 2 CNN blocks and 2 DNN blocks. Each CNN block contains two layers of CNN. The possible adapter is added after each CNN layer, and the adapter in DNN block is plugged after the linear projection layer. Both the decoders for ASR and NLU are AttentionalRNN, which contains 1 RNN block and 1 projection layer. The candidate adapters in ASR and NLU decoders are added behind the RNN block and the projection layer. The language model is not used since the output embedding after softmax in ASR decoder is sent to NLU decoder directly as input for the back-propagation of gradient. 

The loss only considers the NLU loss and the parameter penalty we propose. And NLU CER (character error rate) is chosen as the evaluation metric. The bottleneck adapter (BA) \cite{ba} is selected as the adaptation operation in NFA, which is comprised of a down projection, a non-linearity, and an up-projection, with a skip connection. 
The hidden layer size of BA is set to one quarter of the input size. 
We perform the first-order approximation of Differentiable Architecture Search (DARTS) algorithm \cite{darts} for NAS. The training data is split into two equal parts, one for the architecture parameter learning, and the other for network parameter training. We also experimented other split ratio and the result was not as good as equal dividing. The learning rates for fine-tuning and network parameter learning are set to 0.0001, while the learning rate for architectural parameter learning is 0.001.

\subsection{Results}

\subsubsection{With or Without NAS}

We first experiment the impact of NAS for adapter plugging. The results in Table 1 manifest that using NAS to search appropriate locations for adapters is able to decline the NLU CER significantly. The trainable parameter of NA increases slightly due to the architecture parameter. However, the final selected parameter of NA after the architecture has been learned reduces by 39\% relative to BA. 

\begin{table}[th]
  \caption{The performance of plugging adapters through NAS (NA) or full plugging (FP)}
  \label{tab1}
  \centering
  \begin{tabular}{l l l l l}
    \toprule
    \textbf{Adapt} & \textbf{Total} & \textbf{Train} & \textbf{Selected} & \textbf{NLU} \\
    \textbf{Method} & \textbf{Param.} & \textbf{Param.} & \textbf{Param.} & \textbf{CER} \\
    \midrule
    FP                       & $192.97$M  & $6.04$M & $6.04$M & $30.34\%$          \\
    NA                       & $193.46$M  & $6.52$M & $3.67$M & $17.08\%$             \\
    \bottomrule
  \end{tabular} 
\end{table}
\vspace{-0.5cm}

\subsubsection{NFA v.s. Full Fine-tuning}

Fine-tuning all the parameters in cascaded model (Full Fine-tuning) is the strong baseline for adaptation. The target of our work is to approach the performance of full fine-tuning with less parameters. Plugging adapter in each module and fine-tuning the downstream task is the usual alternative strategy which is parameter efficient. The second row (BA+fine-tuningNLU) in Table 2 demonstrates that the trainable parameters by the method mentioned above scale down substantially while the NLU CER increases. Besides bottleneck adapter, we also test other adapter - gated adapter (GA) \cite{gate_adapt}, which executes an element-wise gated operation on the original network and an expanded adapted weight matrix. The comparison between BA and GA in Table 2 identifies the advantages of BA over GA in the current work. The reason may be the architecture difference between GA and BA, in which BA includes down-sampling and up-sampling operations while GA does not. This may cause a more powerful modeling ability for BA. 

\begin{table}[th]
  \caption{The performance of full fine-tuning and the proposed NFA}
  \label{tab2}
  \centering
  \begin{tabular}{l l l l}
    \toprule
    \textbf{Adapt}& \textbf{Train} & \textbf{Selected} & \textbf{NLU} \\
    \textbf{Method} & \textbf{Param.} & \textbf{Param.} & \textbf{CER} \\
    \midrule
    Full Fine-tuning    & $189.25$M & $189.25$M & $12.42\%$          \\
    BA+fine-tuning NLU    & $14.15$M & $14.15$M & $14.13\%$             \\
    GA+fine-tuning NLU     & $14.17$M & $14.17$M & $18.56\%$             \\
    \midrule
    NFA        & $81.39$M & $65.11$M & $14.57\%$          \\
    ~~+parameter control          & $81.39$M & $16.48$M & $12.81\%$             \\
   ~~~~+two stage            & $81.39$M & $16.48$M & $12.32\%$             \\
    \bottomrule
  \end{tabular}
  
\end{table}

The proposed NFA module expands more trainable parameters due to the fine-tuning operation. However, the final selected parameters declines sharply when the architecture is learned after several epochs, especially after adding the parameter control on loss. More than that, the NLU CER with parameter control is also decreased from 14.57\% to 12.81\%, approaching the full fine-tuning result. If we operate a two-stage training, by which we update both architecture and network parameters alternately in the first stage, while in the second stage only the network parameters are updated, the performance is further enhanced, even surpasses full fine-tuning.

\subsubsection{Ablation}

The amount of trainable parameters for frozen operation in the parameter control loss effects the architecture search significantly and further determines the system performance. We set it as 0 at first. The performance is failed to converge to the full fine-tuning result. We consider that there should be less modules to be frozen for a better transfer, therefore we enlarge the number of $P_{i,fr}$ in the loss penalty and the CER declines successfully. Ultimately, we set $P_{i,fr}$ as half of $P_{i,ft}$, which is in the middle of $P_{i,ad}$ and $P_{i,ft}$, and dynamically changes with different modules. The last line in Table 3 illustrates the dynamic setting is beneficial for the adaptation performance. 

\begin{table}[th]
  \caption{The performance of different settings for frozen operation in the parameter control loss}
  \label{tab3}
  \centering
  \begin{tabular}{l l l}
    \toprule
    \textbf{$P_{i,fr}$} & \textbf{Selected} & \textbf{NLU} \\
     & \textbf{Param.} & \textbf{CER} \\
    \midrule
   $0$    &   $16.01$M & $14.19\%$          \\
  $1000$   &     $18.20$M & $13.87\%$             \\
  Half  &  $16.48$M    &  $12.81\%$  \\
    \bottomrule
  \end{tabular}
  
\end{table}

To investigate the robustness of NFA modules to the number of cascaded tasks, we first cut off SE model in the sequence and only left ASR-NLU. The results in Table 4 verify the robustness of the proposed NFA. The performance of ASR-NLU is inferior to SE-ASR-NLU, indicating the necessity of SE. In addition, we insert a fourth model -wav2vec2- after SE and obtain a similar consistency. The performance degradation of SE-Wav2vec2-ASR-NLU corresponding to SE-ASR-NLU suggests the possible over-fitting due to the large scale of cascaded models.

\begin{table}[th]
  \caption{The performance with different numbers of tasks}
  \label{tab4}
  \centering
  \begin{tabular}{l l l}
    \toprule
    \textbf{Adapt Method} & \textbf{ASR-} & \textbf{SE-Wav2vec2} \\
    & \textbf{NLU} & \textbf{-ASR-NLU} \\
    \midrule
    Full Fine-tuning   &    $13.38\%$ & $15.39\%$          \\
    NFA + parameter control   &     $13.59\%$ & $15.31\%$             \\
    \bottomrule
  \end{tabular}
\end{table}

\vspace{-0.5cm}

\begin{table}[th]
  \caption{The performance with less training data}
  \label{tab5}
  \centering
  \begin{tabular}{l l l l}
    \toprule
    \textbf{Adapt Method} & \textbf{5hr} & \textbf{10hr} & \textbf{20hr} \\
    \midrule
    Full Fine-tuning  &   $27.16\%$  &   $19.50\%$ & $16.69\%$          \\
    NFA + parameter control   &    $24.33\%$  &    $18.90\%$ & $16.25\%$             \\
    +two stage            & &  &              \\
    \bottomrule
  \end{tabular}
  
\end{table}

Finally, we cut down the training data and acquire consistent results (shown in Table 5). Notably, the searched architecture with less training data are identical with the architecture searched by full training set, which supplies a speed-up scheme that we can take less data to gain an optimal architecture and adopt full data to update network parameters. 

\subsection{Analysis}

Figure 3(a) visualizes the searched architecture of the cascaded SE-ASR-NLU framework based on NA. 0 denotes freezing the current module (without an adapter) while 1 stands for plugging an adapter (BA) after the current module. The point line in Figure 3(a) indicates that the selection of whether to add an adapter may be different for modules with the same structure but located in different positions, such as the decoders of ASR and NLU. In addition, the adapters prefer to be located in the top layers of each task and downstream tasks. 

Then we compare the searched results between handcraft fine-tuning (NA+fine-tuningNLU) and the proposed NFA. Figure 3(b) suggests that NFA is able to search reasonable operations for most modules that are similar to manual design.

\vspace{-0.3cm}
\begin{figure}[H]
	\centering  
	\subfigbottomskip=2pt 
	\subfigure[NA]{
		\includegraphics[width=0.48\linewidth]{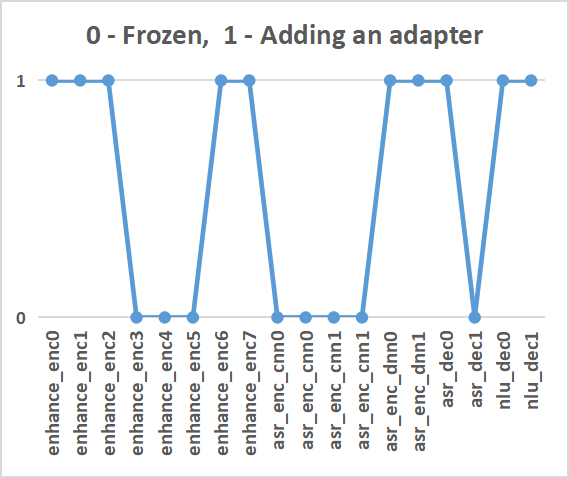}}
	\subfigure[NA+fine-tuningNLU and NFA]{
		\includegraphics[width=0.48\linewidth]{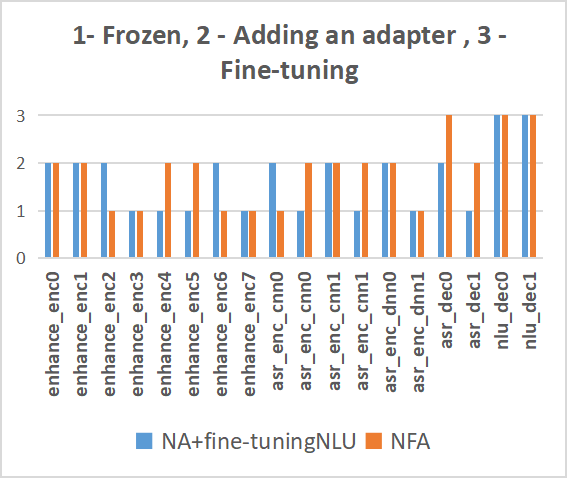}}
  \subfigure[NFA with and without parameter control]{
    \includegraphics[width=0.48\linewidth]{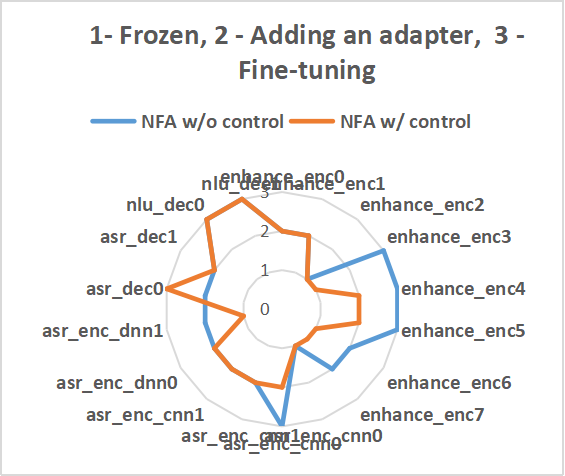}}
  \subfigure[NFA with different adapters]{
    \includegraphics[width=0.48\linewidth]{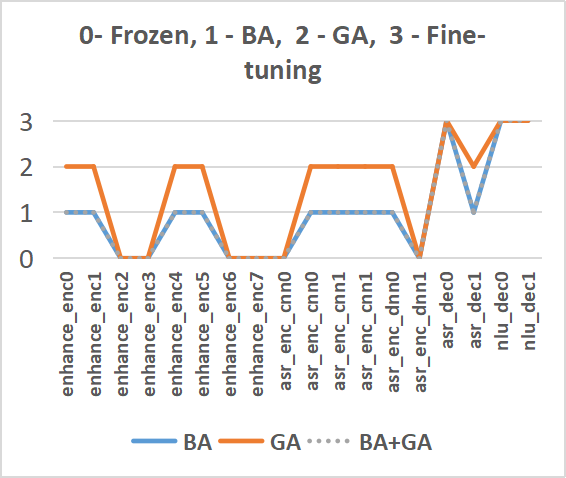}}
	\caption{The searched architecture of the cascaded SE-ASR-NLU framework based on (a) NA, (b) NA+fine-tuningNLU and the proposed NFA, (c) NFA with and without parameter control and (d) NFA with different adapters.}
\end{figure}
\vspace{-0.5cm}

To further validate the effectiveness of the proposed parameter penalty in loss function, we compare the searched architectures of NFA with and without parameter control item in loss function. Figure 3(c) reveals that NFA with parameter control successfully reduces the number of fine-tuned modules.

Moreover, we test adding other type of adapter or more than one type of adapters in NFA. We employ GA as an alternative to BA. The orange line and the blue line in Figure 3(d) indicate that the system has successfully searched consistent operations regardless of adapter type. Besides, the coincidence of the gray line and the blue line illustrates that the system is capable of choosing the optimal adapter when multiple adapters are exploited in NFA.

\section{Conclusion}
In this work, we propose an automatic adaptation scheme for cascaded multi-task systems. We integrate fine-tuning with adapter and frozen as the candidate operations for each adapted module, which simplifies the training procedure and reduces manual intervention. Moreover, we compact the searched architecture through a parameter penalty on the loss. The proposed NFA is able to search reasonable operations for the adaptation of cascaded models and scale down the learnable parameters for efficient optimization without performance degradation.

Future directions include performing NFA in other domains with larger models and combining more tasks.

\bibliographystyle{IEEEtran}
\bibliography{mybib}

\end{document}